\documentclass[11pt,a4paper]{article}
\usepackage[hyperref]{emnlp2020}
\usepackage{times}
\usepackage{latexsym}

\usepackage{microtype}

\aclfinalcopy 
\def\aclpaperid{3211} 

\usepackage{color}
\usepackage{xcolor}
\definecolor{ugreen}{rgb}{0,0.5,0}
\definecolor{lgreen}{rgb}{0.9,1,0.8}
\definecolor{lightgray}{gray}{0.85}
\definecolor{myblack}{rgb}{0.15,0.15,0.15}
\definecolor{lyellow}{rgb}{0.54, 0.25, 0.27}
\usepackage{caption}
\usepackage{enumitem}
\usepackage{amsmath}
\usepackage{amsfonts} 
\usepackage{pgfplots}
\usepackage{tikz}
\usepackage{subfig}
\usetikzlibrary{backgrounds,fit} 
\usetikzlibrary{shapes,arrows,shadows}
\usetikzlibrary{patterns}
\usetikzlibrary{shapes.geometric} 
\usetikzlibrary{decorations.pathreplacing} 
\usetikzlibrary{calc}
\usepackage{array,multirow}
\usepackage{array} 
\usepackage{booktabs} 
\usepackage{arydshln} 
\usepackage{diagbox} 
\newcommand{\PreserveBackslash}[1]{\let\temp=\\#1\let\\=\temp} 
\newcolumntype{C}[1]{>{\PreserveBackslash\centering}p{#1}}
\newcolumntype{R}[1]{>{\PreserveBackslash\raggedleft}p{#1}}
\newcolumntype{L}[1]{>{\PreserveBackslash\raggedright}p{#1}}
\newcolumntype{M}[1]{ >{\centering\arraybackslash}m{#1}}
\usepackage{cleveref}
\crefname{section}{§}{§§}
\Crefname{section}{§}{§§}
\usepackage[ruled,linesnumbered]{algorithm2e}

\pgfplotsset{compat=1.16}

\usepackage{verbatim}

\newlength{\vseg}
\newlength{\hseg}
\newlength{\wnode}
\newlength{\hnode}
\title{Training Flexible Depth Model by Multi-Task Learning \\ for Neural Machine Translation}

\author{
	Qiang Wang$^1$\thanks{\xspace\xspace Work done during Ph.D. study at Northeastern University.},
	Tong Xiao$^{2,3}$
	,
	Jingbo Zhu$^{2,3}$  \\
	$^{1}$Machine Intelligence Technology Lab, Alibaba DAMO Academy\\ 
	$^{2}$Northeastern University, Shenyang, China
	$^{3}$NiuTrans Co., Ltd., Shenyang, China \\ 
	{\tt
		zhiniao.wq@alibaba-inc.com
	}\\
	{\tt
		\{xiaotong,zhujingbo\}@mail.neu.edu.com
	} \\
}

\date{}

\begin{document}
\maketitle

\begin{abstract}
The standard neural machine translation model can only decode with the same depth configuration as training. Restricted by this feature, we have to deploy models of various sizes to maintain the same translation latency, because the hardware conditions on different terminal devices (e.g., mobile phones) may vary greatly. Such individual training leads to increased model maintenance costs and slower model iterations, especially for the industry. In this work, we propose to use multi-task learning to train a flexible depth model that can adapt to different depth configurations during inference. Experimental results show that our approach can simultaneously support decoding in 24 depth configurations and is superior to the individual training and another flexible depth model training method——LayerDrop. 
\end{abstract}

\section{Introduction}


\noindent As neural machine translation models become heavier and heavier \cite{vaswani2017attention}, we have to resort to model compress techniques (e.g., knowledge distillation \cite{hinton2015distilling,kim2016sequence}) to deploy  smaller models in devices with limited resources, such as mobile phones. However, a practical challenge is that the hardware conditions of different devices vary greatly. To ensure the same calculation latency, customizing distinct model sizes (e.g., depth, width) for different devices is necessary, which leads to huge model training and maintenance costs \cite{yu2019slimmable}. For example, we need to distill the pre-trained large model into N individual small models. 
The situation becomes worse for the industry when considering more translation directions and more frequent model iterations.

An ideal solution is to train a single model that can run in different model sizes. Such attempts have been explored in SlimNet \cite{yu2019slimmable} and LayerDrop \cite{fan2020reducing}. SlimNet allows running in four width configurations by joint training of these width networks, while LayerDrop can decode with any depth configuration by applying Dropout \cite{srivastava2014dropout} on layers during training.   
 
\begin{figure*}[t]
	\begin{center}
		\includegraphics[width=0.75\linewidth,height=0.25\linewidth]{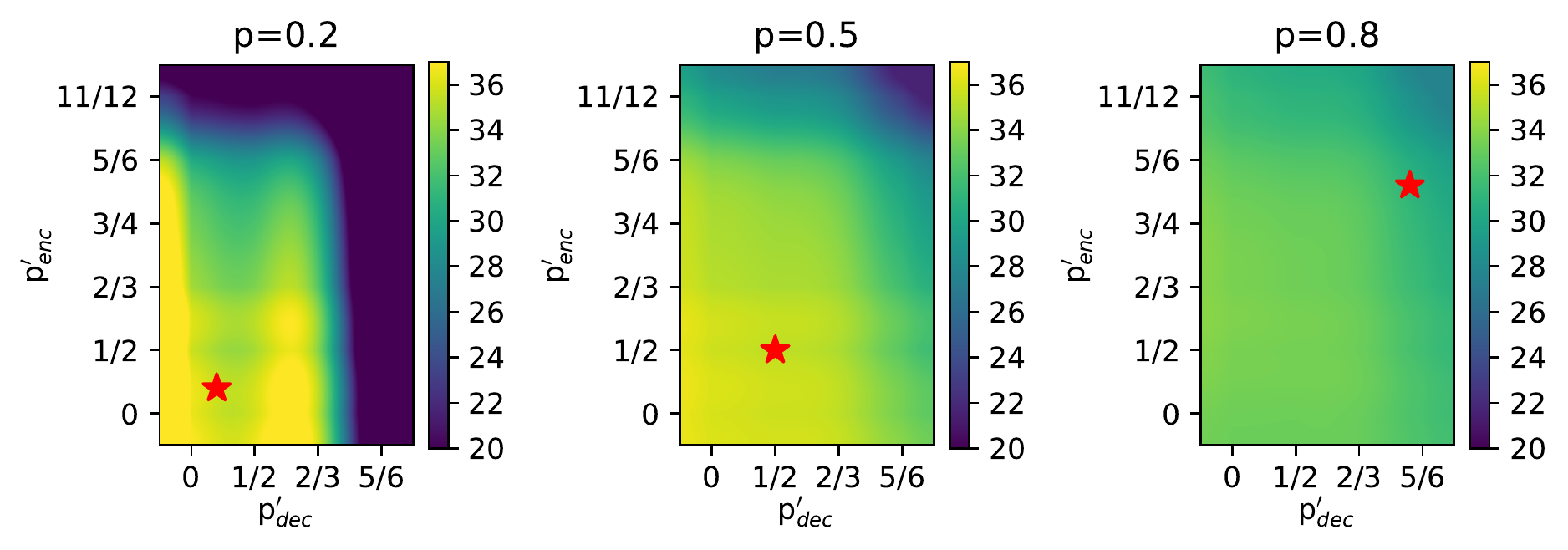}
	\end{center}
	\begin{center}
		\vspace{-0.5em}
		\caption{BLEU score heatmaps of  a 12-layer encoder and a 6-layer decoder model trained by LayerDrop with different layer dropout $p$. p$_{enc}^\prime$ and p$_{dec}^\prime$ denote the layer-prunning ratio at inference on encoder and decoder, respectively. For example, p$_{enc}^\prime$=11/12 means decoding by one encoder layer without the other 11 encoder layers. The red star marks the training layer dropout, i.e. p$_{enc}^\prime$=p$_{dec}^\prime$=p.}
		\label{fig:layerdrop}
		\vspace{-1.em}
	\end{center}
\end{figure*}

In this work, we take a further step along the line of \textit{flexible depth network} like LayerDrop. 
As shown in Figure~\ref{fig:layerdrop}, we first demonstrate that when there is a large gap between the predefined layer dropout during training and the actual pruning ratio during inference, LayerDrop's performance is poor. 
To solve this problem, we propose to use multi-task learning to train a flexible depth model by treating each supported depth configuration as a task. We reduce the supported depth space for the aggressive model compression rate and propose an effective deterministic sub-network assignment method to eliminate the mismatch between training and inference in LayerDrop. 
Experimental results on deep Transformer \cite{wang2019learning} show that our approach can simultaneously support decoding in 24 depth configurations and is superior to the individual training and LayerDrop.

\section{Flexible depth model and LayerDrop}

\subsection{Flexible depth model}
\noindent We first give the definition of \textit{flexible depth model} (FDM): given a neural machine translation model $\mathcal{M}_{M-N}$ whose encoder depth is $M$ and decoder depth is $N$, in addition to (M,N), if $\mathcal{M}_{M-N}$ can also simultaneously decode with different depth configurations $(m_i, n_i)_{i=1}^k$ where $m_i \le M$ and $n_i \le N$ and obtain the comparable performance with independently trained model $\mathcal{M}_{m_i-n_i}$, we refer to $\mathcal{M}_{M-N}$ as a flexible depth model with a capacity of $k$. 
We notice that although a pre-trained vanilla Transformer can force decoding with any depth, its performance is far behind the independently trained model \footnote{BLEU score is only 0.14 if we ask the vanilla Transformer with M=12 and N=6 to decode with M=1 and N=1 directly. However, an individual trained model with M=1 and N=1 can obtain 30.36.  }. Therefore, the vanilla Transformer does not belong to FDM.

\subsection{LayerDrop}

\noindent In NMT, both encoder and decoder are generally composed of multiple layers with residual connections, which can be formally described as:
\begin{equation}
\label{eq:layer}
x_{i+1} = x_i + \textrm{Layer}(x_i).
\end{equation} 

To make the model robust to pruned layers (shallower networks), LayerDrop proposed by \citet{fan2020reducing}, applies structured dropout over layers during training. A Bernoulli distribution associated with a \textit{pre-defined} parameter $p \in$ [0,1] controls the drop rate. It modifies Eq.~\ref{eq:layer} as:
\begin{equation}
\label{eq:layerdrop}
x_{i+1} = x_i + Q_i * \textrm{Layer}(x_i)
\end{equation} 
where $Pr(Q_i=0)=p$ and $Pr(Q_i=1)=1-p$. In this way, the $l$-th layer theoretically can take any proceeding layer as input, rather than just the previous one layer ($l-1$-th layer). 

At runtime, given the desired layer-pruning ratio $p'=1-D_{inf}/D$ where $D_{inf}$ is the number of layers actually used in decoding and $D$ is the total number of layers, LayerDrop selects to remove the $d$-th layer such that:
\begin{equation}
\label{eq:layerdrop_inf}
d \equiv 0(\textrm{mod} \lfloor \frac{1}{p'} \rfloor)
\end{equation} 

\subsection{LayerDrop's problem for flexible depth}
\noindent Although LayerDrop can play a good regularization effect when training deep Transformer \cite{fan2020reducing},  we argue that this method is not suitable for FDM. As illustrated in Figure~\ref{fig:layerdrop}, we demonstrate that LayerDrop suffers a lot when there is a large gap between the pre-defined layer dropout $p$ in training and the actual pruning ratio $p'$ at runtime. 
We attribute it to two aspects:

\begin{enumerate}
	\item \textit{Huge sub-network space in training}. Consider a D-layer network, because each layer can be masked or not, up to $2^D$ sub-networks are accessible during training, which is a major challenge when D is large.
	\item \textit{Mismatch between training and inference.} As opposite to training, LayerDrop uses a deterministic sub-network at inference when given the layer pruning ratio $p'$ (See Eq.~\ref{eq:layerdrop_inf}), which leads to a mismatch between training and inference. For example, for D=6 and D$_{inf}$=3, there are $\tbinom{D}{D_{inf}}$ sub-network candidates during training, while only one of them is used in decoding.
\end{enumerate}

\section{Flexible depth by multi-task learning}
\begin{algorithm}[t]
	\caption{Training Flexible Depth Model by Multi-Task Learning. }  
	\label{alg:mt_train}  
	
	pre-train $\mathcal{M}_{M-N}$ on training data $\mathcal{D}$\;
	generate distillation data $\mathcal{D}'$ by $\mathcal{M}_{M-N}$\;
	$\mathcal{M}'_{M-N} \gets \mathcal{M}_{M-N}$\;
	
	\For {$t$ in $1,2,\ldots,T$} {
		$\mathcal{B} \gets$ sample batch from $\mathcal{D}'$ \;
		gradient $\mathcal{G} \gets 0$\;
		\For { ($m_i$, $n_i$) in $\hat{\phi}(M) \otimes \hat{\phi}(N)$  } {
			
			$\textrm{SN}_{e}$,  $\textrm{SN}_{d}$ $\gets \mathcal{F}(m_i, M)$, $\mathcal{F}(n_i, N)$\;
			Feed $\mathcal{B}$ into network $(\textrm{SN}_{e}, \textrm{SN}_{d})$\;
			Collect gradient $g$ by Back-Propa.\;
			$\mathcal{G} \gets \mathcal{G} + g$\;
		}
		Optimize $\mathcal{M}'_{M-N}$ with gradient $\mathcal{G}$\;
	}	
	Return $\mathcal{M}'_{M-N}$
\end{algorithm}

We propose to use multi-task learning to solve the above problems. All tasks are trained jointly and share the same parameters. 
Concretely, unlike LayerDrop, which allows up to $M \times N$ possible depth configurations, our approach sets a smaller depth configuration space $(m_i, n_i)_{i=1}^k (k<M \times N)$ in advance and takes each $(m_i, n_i)$ as a task. 
Another major difference from LayerDrop is that each task's sub-network is unique and deterministic in our method, resulting in consistent sub-network used between training and inference.


\begin{enumerate}[leftmargin=0pt]
	
	\item[] \textbf{Reduce depth space. }  For depth D, in principle, LayerDrop can be pruned to any depth of $\phi(D)=\{0, 1, 2, \ldots, D\}$. However, consider the actual situation of model compression for resource-limited devices, it is unnecessary if the compressing rate is too low, e.g., $D$ $\rightarrow$ $D$-1. Therefore, for an aggressive compress rate, we replace the entire space $\phi(D)$ with the set of all positive divisors of $D$ \footnote{For the diversity of depth configuration, we assume that $D$ is not a prime number in this work. }:
	\begin{equation}
		\label{eq:mt_space}
		\hat{\phi}(D)=\{d|D\%d=0, 1 \le d \le D\}
	\end{equation} 
	The physical meaning of $\hat{\phi}(D)$ is to compress every $D/d$ layers into one layer, where $d \in$ $\hat{\phi}(D)$. 
	
	\item[] \textbf{Guideline for deterministic sub-network assignment. }
	The use of deterministic sub-networks is critical to maintaining the consistency between training and inference. 
	However, for each $d \in \hat{\phi}(D)$, it is not trivial to decide which $d$ layers should be selected to construct the $d$-layer sub-network.
	Here we propose two metrics to guide the procedure. The first is \texttt{task balance} (TB), whose motivation is to make every layer have as uniform tasks as possible. We use the standard deviation of the number of tasks per layer to measure it quantitatively:
	\begin{equation}
		\label{eq:bt}
		\textrm{TB} = \sqrt{\frac{\sum_{i \in [1,D]}{\Big(t(i) - \bar{t}\Big)^2}}{D}} 
	\end{equation}
	where $t(i)$ is the number of tasks in which the $i$-th layer participates and $\bar{t}=\frac{\sum_{d \in \hat{\phi}(D)}{d}}{D}$. 
	The second is \textit{average layer distance} (ALD), which requires the distance between adjacent layers in the sub-network $\textrm{SN}(d)=\{L_{a_1}, L_{a_2}, \ldots, L_{a_d}\}$ should be large. For example, for a 6-layer network, if we want to build a 2-layer sub-network, it is unreasonable to select $\{L_1, L_2\}$ directly because the features extracted by adjacent layers are semantically similar \cite{peters2018deep,raganato-tiedemann-2018-analysis}. Therefore, we use the average distance between layers in all sub-networks as the metric:
	\begin{equation}
		\label{eq:ald}
		\textrm{ALD} = \frac{\sum \limits_{d \in \hat{\phi}(D)}{\sum \limits_{a_i, a_{i+1} \in \textrm{SN}(d)}{|a_{i+1} - a_{i}|}}}{Z}  
	\end{equation}
	where $Z=\sum_{d \in \hat{\phi}(D)}{(d-1)}$ is the normalization item.
	
	\item[] \textbf{Proposed method. }  
	Guided by these two metrics, we design an effective sub-network assignment method \texttt{Optimal}. 
	We record the usage state $s_i$ of each layer to ensure not to put too many tasks on the same layer. At initialization, we set $s_i$ as \textit{Alive}. For $d \in \hat{\phi}(D)$, 
	\texttt{Optimal} prioritizes to process large depth. \texttt{Optimal} uniformly assigns one layer for every $c=D/d$ layers to make ALD high. 
	In each chunk, we pick the middle layer of $\textrm{ceil}(c/2)-1$ (called \texttt{MiddleLeft}). Note that, LayerDrop uses the leftmost layer in each chunk (called \texttt{Left}), as shown in Eq.~\ref{eq:layerdrop_inf}. Although \texttt{Left} and \texttt{MiddleLeft} have the same ALD, we found that there is a large gap in TB. For example, when $D$=12, \texttt{Left}'s TB is 1.5, which is much higher than \texttt{MiddleLeft}'s 0.78 (lower is better). Then, \texttt{Optimal} records which layers are used and picks the less used layers as much as possible. Each used layer is marked as \textit{Dead}. If current alive layers cannot accommodate the picked depth $d$, we pass it and choose a smaller $d$ until the alive layers are sufficient, or reset all layers as \textit{Alive}.

	\begin{table*}[t]
		\begin{center}
			\setlength{\tabcolsep}{5pt}
			\resizebox{0.9\textwidth}{!}
			{
				\begin{tabular}{c| ccc |ccc |ccc |ccc}
					\toprule[1pt]
					
					\multicolumn{1}{c|}{\multirow{2}{*}{\diagbox{M}{N}}} 
					& \multicolumn{3}{c |}{1} &
					\multicolumn{3}{c |}{2} &
					\multicolumn{3}{c |}{3} &
					\multicolumn{3}{c }{6} \\
					
					~ & Base & $\Delta_{LD}$ & $\Delta_{MT}$ & Base & $\Delta_{LD}$ & $\Delta_{MT}$ & Base & $\Delta_{LD}$ & $\Delta_{MT}$ & Base & $\Delta_{LD}$ & $\Delta_{MT}$ \\
					\hline
					
					1 & \textbf{31.54} & -3.04 & -0.09 & 33.38 & -2.37 & \textbf{+2.67} & 33.87 & -1.99 & \textbf{+0.64} & \textbf{34.77} & -2.27 & -0.03 \\
					2 & 32.80 & -0.98 & \textbf{+0.31} & 34.15 & -0.53 & \textbf{+0.48} & 
					34.58 & -0.22 & \textbf{+0.55} & 34.95 & -0.15 & \textbf{+0.49} \\
					
					3 & 33.38 & -0.40 & \textbf{+0.26} & 34.40 & +0.15 & \textbf{+0.65} & 34.74 & +0.40 & \textbf{+0.75} & 35.29 & +0.25 & \textbf{+0.52} \\
					
					4 & 33.92 & -0.27 & \textbf{+0.44} & 34.77 & +0.38 & \textbf{+0.59} & 35.01 & +0.50 & \textbf{+0.86} & 35.37 & +0.41 & \textbf{+0.68} \\
					
					6 & 34.28 & -0.29 & \textbf{+0.07} & 35.06 & +0.20 & \textbf{+0.42} & 35.23 & +0.41 & \textbf{+0.61} & 35.51 & 0.34 & \textbf{+0.51} \\
					
					12 & 34.72 & -0.05 & \textbf{+0.06} & 35.26 & +0.49 & \textbf{+0.53} & 35.52 & \textbf{+0.53} & +0.44 & 35.74 & \textbf{+0.49} & +0.48 \\
					\bottomrule[1pt]
				\end{tabular}
			}

			\vspace{-0.5em}
			\caption{BLEU scores of \texttt{Baseline}/\texttt{LayerDrop}/\texttt{MT} in all tasks (6$\times$4). 
				$\Delta_{LD}$/$\Delta_{MT}$ represents the BLEU score difference between \texttt{LayerDrop}/\texttt{MT} and \texttt{Baseline}, respectively.
				All the three methods have the same training cost. 
				Boldface denotes the winner.}
			\label{table:main_results}
			\vspace{-1.em}
		\end{center}
	\end{table*}

	\item[] \textbf{Training. }   
	Algorithm~\ref{alg:mt_train} describes the training process of our method. During training, compared with individual training and LayerDrop from scratch, our FDM finetunes on the individually pre-trained $\mathcal{M}_{M-N}$ and uses sequence-level knowledge distillation (Seq-KD) \cite{kim2016sequence} to help shallower networks training. We note that in conventional Seq-KD, the student model cannot finetune on the teacher model directly because the two models have different sizes. However, FDM allows models with different depths to share the same parameters, and finetuning on the pre-trained teacher model also promotes model convergence. 
	
\end{enumerate}

\section{Experiments}

\subsection{Setup}
\begin{table}[t]
	\centering
	\resizebox{0.9\linewidth}{!}
	{
		\begin{tabular}{c  c c}
			\toprule[1pt]
			
			\textbf{System} & \multicolumn{1}{c }{\textbf{w/o Seq-KD}} &
			\multicolumn{1}{c }{\textbf{w/ Seq-KD}} \\
			\hline 
			
			Baseline & 33.92 & 34.51 \\
			LayerDrop & 32.80 & 34.18 \\
			MT & 34.07 & 34.95 \\	 
			\bottomrule[1pt]
		\end{tabular}	
	}
	
	\vspace{-0.5em}
	\caption{Average BLEU scores of 24 tasks on test set w.r.t. Seq-KD.}
	\label{table:seqkd}
	\vspace{-1.em}
\end{table}

\noindent We conducted experiments on IWSLT'14 German$\rightarrow$English (De$\rightarrow$En, 160k) following the same setup as \citet{wu2019pay}. 
To verify FDM's efficiency, we train all models with a deep encoder to contain more tasks. Specifically, we train a PreNorm Transformer \cite{wang2019learning}  with M=12 and N=6. See Appendix~A for the details. 


We mainly compare our method \texttt{MT} with the two baselines: \texttt{Baseline} and \texttt{LayerDrop}.
\texttt{Baseline} denotes individually training the standard Transformer from scratch with different depths.
For fair comparisons, both \texttt{Baseline} and \texttt{LayerDrop} use Seq-KD during training and have the same training costs \footnote{Original LayerDrop in \citet{fan2020reducing} samples a batch to update the model, while we modify it by accumulating 6$\times$4=24 batches to keep the training cost comparable with \texttt{Baseline} and \texttt{MT}. Also, more samples improve LayerDrop's performance. For example, the average BLEU score in 24 tasks with one batch and 24 batches is 32.31 and 34.18, respectively. }.


\subsection{Results and Analysis}

\begin{enumerate}[leftmargin=0pt]
	
	\item[] \textbf{Main results. } As shown in Table~\ref{table:main_results}, we compared \texttt{Baseline}, \texttt{LayerDrop} and our \texttt{MT} in all tasks. Although \texttt{LayerDrop} outperforms our method when a few layers pruned, we can see that \texttt{MT} is the winner in most tasks (20/24). It indicates that our method is superior to LayerDrop for FDM training and demonstrates the potential to substitute a dozen models with different depths to just one model. Besides, in line with \citet{fan2020reducing}, it is interesting to see the FDM without any pruning outperforms the individually trained model (see M=12, N=6), which is obvious evidence that jointly training of various depth models has a good regularization effect.
	
	\item[] \textbf{Knowledge distillation. } Table~\ref{table:seqkd} shows average BLEU scores of 24 tasks when training a flexible depth model with/without Seq-KD. 
	It is clear that using distillation data helps FDM training in all systems, which is in line with the previous single-model compression study \cite{kim2016sequence}. According to \citet{zhou2020understanding}, Seq-KD makes the training data distribution smoother, so we suspect that FDM benefits from Seq-KD because of the difficulty of multi-task learning. 
	
	\begin{table}[t]
		\begin{center}
		
		\resizebox{0.9\linewidth}{!}
		{
			\begin{tabular}{c  c c c}
				\toprule[1pt]
				
				\textbf{Strategy} & \multicolumn{1}{c }{\textbf{TB}$\downarrow$} &
				\multicolumn{1}{c }{\textbf{ALD}$\uparrow$} & \multicolumn{1}{c }{\textbf{BLEU$_{6 \times 4}$}} \\
				\hline 
				
				Head & 1.78 & 1.0 & 34.37 \\
				Seq & 0.49 & 1.0 & 34.53 \\
				Left & 1.50 & 2.0 & 34.59 \\
				MiddleLeft & 0.78 & 2.0 & 34.90 \\
				Optimal & 0.49 & 2.05 & 34.95 \\
				\bottomrule[1pt]
			\end{tabular}		
		}
		
		\end{center}
		
		\vspace{-.5em}
		\caption{Average BLEU scores of 24 tasks on test set w.r.t. sub-network strategy. We report TB and ALD on encoder side. $\downarrow$ denotes the lower the better, while $\uparrow$ is on contrary. Note that, unlike the standard BLEU score, $\textrm{BLEU}_{6\times4}$ is more difficult to change significantly because it is scaled of the number of tasks.}
		\label{table:strategy}
		\vspace{0em}
	\end{table}
	
	\item[] \textbf{Sub-layer assigiment strategy. } 
	Besides the proposed \texttt{Optimal} and \texttt{Left} used by LayerDrop and its improved version \texttt{MiddleLeft}, we also compared with the other two strategies:  \texttt{Head} and \texttt{Seq}, to check the consistency between BLEU and the proposed guidelines (TB and ALD).
	\texttt{Head} is the simplest method, which always picks the first $d$ layers as the sub-network. However, it causes the bottom layers heavier than the top layers. \texttt{Seq} avoids this problem by sequentially skipping previously used layers. For example, for $D$=6, $d$=1, \texttt{Seq} first uses $L_1$ as the sub-network. Next, when $d=2$, \texttt{Seq} selects $L_2$ and $L_3$. This method ensures that the minimal burden on all layers, but it violates the ALD metrics.
	Table~\ref{table:strategy} shows the average BLEU scores on all tasks by several sub-network strategies. 
	While \texttt{MiddleLeft} already has good TB and ADL, we argue that it is not the best. This is because \texttt{MiddleLeft} treats each $d$ independently regardless of which layers are used in the previous $d'$.
	We can see the proposed policy with lower TB and higher ALD obtains the best result, which indicates that our proposed metrics are helpful to determine which strategy is sound.

	\begin{table}[t]
		\centering
			\resizebox{\linewidth}{!}
			{
				\begin{tabular}{c  c c c}
					\toprule[1pt]
					
					\textbf{System} & \multicolumn{1}{c }{\textbf{\# task}} & \multicolumn{1}{c }{\textbf{$\textrm{BLEU}_{N=6}$}} &
					\multicolumn{1}{c }{\textbf{$\textrm{BLEU}_{M=12}$}} \\
					\hline 
					
					Baseline & 1 & 35.27 & 35.31 \\
					MT (only encoder) & 6 & 35.79 & N/A \\
					MT (only decoder)  & 4 & N/A & 35.80 \\
					MT (both) & 24 & 35.71 & 35.68 \\
					\bottomrule[1pt]
				\end{tabular}
			}
		
		\vspace{-0.5em}
		\caption{Average BLEU scores when reducing the number of tasks.}
		\label{table:simpler_task}
		\vspace{-1.em}
	\end{table}
	
	\item[] \textbf{Reduce the number of tasks. }  
	Intuitively, the number of tasks demines the learning difficulty of our method. To verify this assumption, we tested the other two baselines: (1) only training the flexible-depth encoder (depth from \{1, 2, 3, 4, 6, 12\}) but the decoder depth is the constant 6, denoted by \textit{MT (only encoder)}; (2) only training the flexible-depth decoder (depth from \{1, 2, 3, 6\}) but the encoder depth is the constant 12, denoted by \textit{MT (only decoder)}.  Then we compared the average BLEU scores under fixing the decoder depth as 6 ($\textrm{BLEU}_{N=6}$) and fixing the encoder depth as 12 ($\textrm{BLEU}_{M=12}$).
	As shown in Table~\ref{table:simpler_task}, when we reduce the number of tasks, we can generally obtain better performance. It indicates that if removing some unnecessary tasks, our FDM has the potential for further improvement.
	
	\begin{table}[t]
		\begin{center}
		\resizebox{0.9\linewidth}{!}
		{
				\begin{tabular}{c c  c c c}
				\toprule[1pt]
				
				\textbf{Burden} & \multicolumn{1}{c }{\textbf{Batch}} &
				\multicolumn{1}{c }{\textbf{\#Enc.}} & \multicolumn{1}{c }{\textbf{\#Dec.}} & \multicolumn{1}{c }{\textbf{BLEU}$_{6 \times 4}$} \\
				\hline 
				
				\multicolumn{1}{l|}{100\%} & $B$ & 6 & 4 & 34.95 \\
				\hline
				
				\multicolumn{1}{l|}{\multirow{3}{*}{50\%}} &
				$B$ & 6 & 2 & \textbf{34.79} \\
				\multicolumn{1}{l|}{} & $B$ & 3 & 4 & 34.74 \\
				\multicolumn{1}{l|}{} & 0.5$B$ & 6 & 4 & 34.77 \\
				\hline
				
				\multicolumn{1}{l|}{\multirow{5}{*}{25\%}} &
				$B$ & 6 & 1 & \textbf{34.67} \\
				\multicolumn{1}{l|}{} & $B$ & 3 & 2 & 34.58 \\
				\multicolumn{1}{l|}{} & 0.5$B$ & 6 & 2 & 34.51 \\
				\multicolumn{1}{l|}{} & 0.5$B$ & 3 & 4 & 34.54 \\
				\multicolumn{1}{l|}{} & 0.25$B$ & 6 & 4 & 34.52 \\
				
				\bottomrule[1pt]
			\end{tabular}	
		}
		
		\end{center}
		
		\begin{center}
			\vspace{-0.5em}
			\caption{BLEU scores against training efficiency. $B$ denotes the full token-level batch size of 8k. $\textrm{BLEU}_{6\times4}$ represents the average BLEU scores on 24 tasks. }
			\label{table:sampling_efficient}
			\vspace{-1.em}	
		\end{center}
		
	\end{table}

	\item[] \textbf{Training efficiency. } 
	Our multi-task learning needs to accumulate gradients on all tasks, and its cost is linearly related to the number of tasks.
	Actually, we can sample fewer tasks instead of enumerating them all. For example, randomly sampling 3 tasks from 6 depth candidates (denoted by \textbf{\#Enc.=3}). Another way to reduce training costs is to use smaller batches. We compared different strategies at \{100\%, 50\%, 25\%\} training costs, as shown in Table~\ref{table:sampling_efficient}. First of all, we can see that more training costs can obtain better performance. Compared with reducing tasks and reducing batches, we found that the former is a better choice. In particular, sampling more depths on the encoder side is more important than the decoder side, which is consistent with the recent observation in \citet{wang2019learning} that encoder is more important than decoder in terms of translation performance.

\end{enumerate}

\section{Conclusion}
\label{sec:conclusion}
We demonstrated LayerDrop is not suitable for FDM training because of (1) the huge sub-network space in training and (2) the mismatch between training and inference. Then we proposed to use multi-task learning to mitigate it. Experimental results show that our approach can decode with up to 24 depth configurations and obtain comparable or better performance than individual training and LayerDrop. In the future, we plan to explore more effective FDM training methods, and combining flexible depth and width is also one of the attractive directions.

\section*{Acknowledgements}
This work was supported in part by the National Science Foundation of China (Nos. 61876035 and 61732005), the National Key R\&D Program of China (No. 2019QY1801).

\bibliography{emnlp2020}
\bibliographystyle{acl_natbib}

\clearpage
\appendix

\section{Detailed training setup}
\label{sec:proof}

\noindent We replicate the model configuration (embed=512, ffn=1024, head=4) as the baseline in (Wu et al., 2019). In addition, the batch size of 8192, attention dropout of 0.1, and relu dropout of 0.1 is used as suggested by \citet{wang2019learning}. When training the standard Transformer from scratch, we follow the \textit{inverse\_sqrt} learning rate schedule with learning rate of 0.0015 and warmup of 8k. To speed up convergence of FDMs (e.g. \texttt{MT}, \texttt{LayerDrop}), all of them are finetuned from the pre-trained baseline with learning rate of 0.0005 and warmup of 4k.

At inference, we use a beam size of 5 and average last 5 checkpoints. We use case insensitive BLEU score evaluated by \textit{multi-bleu.perl} as previous works. 

\end{document}